\documentclass[11pt]{article}

\usepackage[final]{acl}

\usepackage{times}
\usepackage{latexsym}

\usepackage[T1]{fontenc}
\usepackage[utf8]{inputenc}

\usepackage{microtype}

\usepackage{inconsolata}

\usepackage{graphicx}

\usepackage{booktabs}   % For professional-quality tables
\usepackage{amsmath}    % For mathematical formatting
\usepackage{subcaption} % For side-by-side plots if needed
\usepackage{comment}   % Fixes "Environment comment undefined"
\usepackage{multirow}  % Fixes "\multirow" errors
\usepackage{amssymb}   % Fixes "\mathbb" errors
\usepackage{amsmath}   % Recommended for complex math
\usepackage{tabularx}
\usepackage{booktabs}

\title{The Silent Vote: Improving Zero-Shot LLM Reliability by Aggregating Semantic Neighborhoods}
\author{Sanket Badhe \\
  \texttt{sanketbadhe@google.com} \\\And
  Priyanka Tiwari \\
  \texttt{tiwarip@alumni.purdue.edu} \\\And
  Deep Shah \\
  \texttt{shahdeep@google.com}}

\begin{document}
\maketitle
\begin{abstract}
Large Language Models are increasingly used as zero-shot classifiers in complex reasoning tasks. However, standard constrained decoding suffers from a phenomenon we define as Renormalization Bias. When a model is restricted to a small set of target labels, the standard softmax operation discards the probability mass assigned to semantic synonyms in the original distribution. This loss of information, which we call the Silent Vote, results in artificial overconfidence and poor calibration. 

We propose Semantic Softmax, an inference-time layer that recovers this lost information by aggregating the scores of the semantic neighborhood surrounding each target label. We evaluate this approach on Qwen-3 and Phi-4-mini models using GoEmotions and Civil Comments datasets. Our results demonstrate consistent improvements across all evaluation metrics: Semantic Softmax substantially reduces Expected Calibration Error (ECE) and Brier Score, while simultaneously enhancing discriminative performance in terms of AUROC and Macro-F1. By accounting for linguistic nuances, our method provides a more calibrated and accurate alternative for zero-shot classification.

\end{abstract}

\section{Introduction}
Large Language Models (LLMs) have shifted the paradigm of text classification from supervised training to zero-shot inference \citep{NEURIPS2020_1457c0d6}. In the LLM as a Classifier framework, models are prompted to select a label from a predefined set of candidates \citep{doi:10.1073/pnas.2305016120, badhe2026promptlevel}. To ensure the model adheres to these task-specific constraints, practitioners typically employ constrained decoding \citep{shin-etal-2021-constrained}. To enforce these task-specific constraints, practitioners typically compute probabilities by restricting normalization to the target label tokens, effectively discarding probability mass assigned to all other vocabulary items \citep{schick-schutze-2021-exploiting}.

While effective for enforcing output formats, this methodology introduces a systematic distortion defined here as Renormalization Bias. Standard constrained decoding operates under a localized assumption that all evidence for a specific class is contained within its representative token. However, linguistic meaning in LLMs is inherently distributed across a dense semantic neighborhood of synonyms and related concepts \citep{NIPS2013_9aa42b31}. When a model predicts a text is thrilling but is forced to choose between the labels Joy or Sadness, the probability mass assigned to those synonyms is discarded before the final distribution is calculated.

This exclusion results in a systematic calibration error. By filtering out the probability density assigned to synonymous tokens, the standard constrained softmax operation systematically overestimates the posterior probability of the remaining targets. This leads to severe miscalibration, a known issue where instruction-tuned models exhibit extreme overconfidence \citep{pmlr-v70-guo17a, NEURIPS2021_8420d359}. This results in a model that exhibits high-confidence miscalibration. The system may assign near-certain probability to a target label solely because it is the only permissible token with non-negligible affinity to the underlying prediction, even if the absolute semantic alignment is weak. Such behavior prevents the model from accurately quantifying the aleatoric uncertainty inherent in ambiguous inputs \citep{nie-etal-2020-adversarial, pavlopoulos-etal-2020-toxicity}.

 To address these operational limitations, this work proposes Semantic Softmax, a lightweight, training-free inference layer designed to recover this \textit{Silent Vote}. Instead of blindly masking the vocabulary, Semantic Softmax utilizes the internal output embeddings of the model to aggregate probability mass from the entire semantic neighborhood surrounding each target label. Because it requires no parameter updates or structural model changes, it functions as a highly deployable, plug-and-play solution for existing serving infrastructure.

The contributions of this paper are three-fold: 
\begin{enumerate}
    \item We formalize the concept of Renormalization Bias and its negative impact on the reliability of enterprise LLM deployments.
    \item We introduce Semantic Softmax, an operationally feasible and mathematically grounded method to aggregate distributed semantic evidence during constrained inference.
    \item Through experiments on GoEmotions \citep{demszky-etal-2020-goemotions} and and Civil Comments \citep{pavlopoulos-etal-2020-toxicity}, we demonstrate that our method drastically reduces Expected Calibration Error (ECE) and Brier Score. It improves alignment with human disagreement in ambiguous scenarios while simultaneously enhancing discriminative performance in terms of AUROC and Macro-F1.
\end{enumerate}

\section{Related Work}
The reliability of LLMs in zero-shot classification depends on the alignment between internal representations and task constraints.

\subsection{Constrained Inference and Decoding}
Constrained decoding is the standard mechanism for enforcing output formats in generative models \citep{shin-etal-2021-constrained}. Techniques such as verbalizer-based logit masking allow practitioners to restrict the vocabulary to valid labels \citep{gao-etal-2021-making}. Research into guided generation has focused largely on maintaining structural adherence or satisfying schema requirements \citep{willard2023efficientguidedgenerationlarge}, whereas this paper investigates the negative impact of these constraints on probability calibration.

\subsection{Calibration in Large Language Models}
Calibration measures the degree to which a predicted probability reflects actual accuracy \citep{pmlr-v70-guo17a}. Studies have shown that while LLMs demonstrate high zero-shot performance, they are frequently miscalibrated, often exhibiting extreme overconfidence \citep{NEURIPS2021_8420d359, kadavath2022languagemodelsmostlyknow}. This issue is exacerbated by alignment techniques like RLHF, which push models toward more peaked distributions and lower output diversity \citep{kirk2024understanding, tian-etal-2023-just}. We identify Renormalization Bias as a significant driver of this overconfidence.

\subsection{Uncertainty and Ambiguity in NLP}
Human language is inherently ambiguous, leading to significant disagreement among annotators \citep{nie-etal-2020-adversarial, kath2026large}. This aleatoric uncertainty is a critical signal for reliable systems \citep{pavlopoulos-etal-2020-toxicity}. Standard pipelines often collapse this uncertainty by forcing a single label choice. Semantic Softmax provides an inference-time method to recover human-like uncertainty without additional training.

\section{Methodology and Experimental Setup}

This section formalizes the theoretical framework of Renormalization Bias and details the proposed Semantic Softmax intervention. We further outline the datasets, model configurations, and evaluation metrics used to validate the efficacy of our approach.

\subsection{Formalizing Renormalization Bias and the Synonym Trap}

In the standard paradigm of LLM classification, a model is prompted to produce a single token representing a label from a discrete candidate set $\mathcal{L} = \{l_1, l_2, \dots, l_n\}$. Given a prompt $x$, the model generates a raw logit vector $\mathbf{z}$ over the entire vocabulary $\mathcal{V}$. To enforce task constraints, practitioners apply a mask $M$, where $M_i = 0$ if $i \in \mathcal{L}$ and $M_i = -\infty$ otherwise. The resulting constrained probability distribution is calculated via a standard softmax operation:

\begin{equation}
P(l_j | x) = \frac{\exp(z_{l_j})}{\sum_{l' \in \mathcal{L}} \exp(z_{l'})}
\end{equation}

This localized approach assumes that the semantic evidence for a specific category is concentrated entirely within the single token representing that class. However, instruction-tuned models learn distributed representations where meaning is spread across a dense neighborhood of synonyms and related concepts \citep{NIPS2013_9aa42b31}. If a model assigns significant probability mass to related tokens $v \notin \mathcal{L}$ such as synonyms or hyponyms of $l_j$, this mass is discarded during the renormalization process. 

We define this loss of information as Renormalization Bias. The exclusion of these related tokens creates a Synonym Trap where the model is forced to ignore the Silent Vote of its broader vocabulary. Consequently, the resulting distribution becomes artificially peaked, leading to the extreme overconfidence and miscalibration frequently observed in zero-shot classifiers \citep{pmlr-v70-guo17a, NEURIPS2021_8420d359}.

\subsection{The Semantic Softmax Framework}

To mitigate Renormalization Bias, we propose Semantic Softmax, a method that aggregates the distributed semantic evidence before probability calculation. Instead of masking the vocabulary, we utilize the top $K$ tokens from the unconstrained distribution, denoted as $\mathcal{V}_{topK}$. We define a Semantic Kernel that measures the relationship between these vocabulary tokens and our target labels using the model output embeddings $\mathbf{E} \in \mathbb{R}^{|\mathcal{V}| \times d}$.

The semantic weight $w(v, l)$ between an unconstrained vocabulary token $v$ and a target label $l$ is calculated using a thresholded cosine similarity:

\begin{equation}
w(v, l) = \max\left(0, \frac{\mathbf{E}_v \cdot \mathbf{E}_l}{\|\mathbf{E}_v\| \|\mathbf{E}_l\|} - \tau\right)
\end{equation}

In this formulation, $\tau$ is a hyperparameter threshold that serves as a noise filter to ensure only semantically relevant tokens contribute to the final score. The final probability for each target label is then computed as a weighted sum of the probabilities of the top $K$ tokens:

\begin{equation}
P_{sem}(l | x) = \frac{\sum_{v \in \mathcal{V}_{topK}} P(v) \cdot w(v, l)}{\sum_{l' \in \mathcal{L}} \sum_{v \in \mathcal{V}_{topK}} P(v) \cdot w(v, l')}
\end{equation}

By aggregating mass from the semantic neighborhood, Semantic Softmax allows the model to reflect its internal aleatoric uncertainty. If the model logic is distributed across tokens associated with different target classes, the resulting distribution will naturally soften, thereby improving calibration.

\subsection{Data and Evaluation Protocol}

\textbf{Model Configuration:} We evaluate on Qwen-3-1.7B and Phi-4-mini (3.8B). This setup confirms that Renormalization Bias is a structural decoding artifact independent of specific model. For all experiments, we maintain a top-k value of $K=50$ and a similarity threshold of $\tau=0.8$.

\textbf{Benchmark Datasets:}
\begin{enumerate}
    \item \textbf{GoEmotions} \citep{demszky-etal-2020-goemotions}: This dataset consists of 58,000 Reddit comments labeled with 28 fine-grained emotion categories. It serves as a primary testbed for synonym recovery due to the heavy semantic overlap between labels such as admiration, approval, and pride.
    \item \textbf{Civil Comments (Ambiguous Subset)}: This dataset is used to evaluate the model's ability to handle aleatoric uncertainty. Unlike datasets that force a hard label, Civil Comments provides scores derived from annotator disagreement, representing a consensus mean. \citep{pavlopoulos-etal-2020-toxicity}.
\end{enumerate}

\textbf{Evaluation Metrics:}
To provide a comprehensive view of model reliability, we utilize the following metrics:
\begin{itemize}
    \item \textbf{Expected Calibration Error (ECE)}: This metric quantifies the average gap between the model's predicted confidence and its actual accuracy across different confidence bins.
    \item \textbf{Brier Score}: A proper scoring rule that measures the mean squared difference between predicted probabilities and actual outcomes, serving as a robust indicator of calibration.
    \item \textbf{Macro-F1 and AUROC}: These metrics ensure that the semantic aggregation preserves the discriminative power of the model and does not trade off accuracy for calibration.
\end{itemize}

\begin{table*}[t]
\centering
\small
\begin{tabular}{lllcccc}
\toprule
\textbf{Model} & \textbf{Dataset} & \textbf{Method} & \textbf{ECE $\downarrow$} & \textbf{Brier $\downarrow$} & \textbf{AUROC $\uparrow$} & \textbf{F1 $\uparrow$} \\
\midrule
\multirow{4}{*}{Qwen-3-1.7B} & \multirow{2}{*}{GoEmotions} & Standard & 0.574 & 0.842 & 0.712 & 0.229 \\
& & \textbf{Semantic (Ours)} & \textbf{0.069} & \textbf{0.591} & \textbf{0.763} & \textbf{0.267} \\
\cmidrule{2-7}
& \multirow{2}{*}{CivilComments} & Standard & 0.482 & 0.571 & 0.784 & 0.412 \\
& & \textbf{Semantic (Ours)} & \textbf{0.108} & \textbf{0.517} & \textbf{0.882} & \textbf{0.451} \\
\midrule
\multirow{4}{*}{Phi-4-mini} & \multirow{2}{*}{GoEmotions} & Standard & 0.421 & 0.795 & 0.744 & 0.236 \\
& & \textbf{Semantic (Ours)} & \textbf{0.065} & \textbf{0.588} & \textbf{0.756} & \textbf{0.253} \\
\cmidrule{2-7}
& \multirow{2}{*}{CivilComments} & Standard & 0.395 & 0.542 & 0.812 & 0.421 \\
& & \textbf{Semantic (Ours)} & \textbf{0.092} & \textbf{0.498} & \textbf{0.835} & \textbf{0.451} \\
\bottomrule
\end{tabular}
\caption{Comparative performance of Standard vs. Semantic Softmax across both Qwen and Phi models. Semantic Softmax consistently reduces ECE and Brier Score across both GoEmotions and CivilComments while simultaneously enhancing discriminative power (AUROC and F1). }
\label{tab:main_results}
\end{table*}

\begin{figure*}[t]
    \centering
    % Placeholder for your 2-panel figure (generated by the Python script)
    \includegraphics[width=\textwidth]{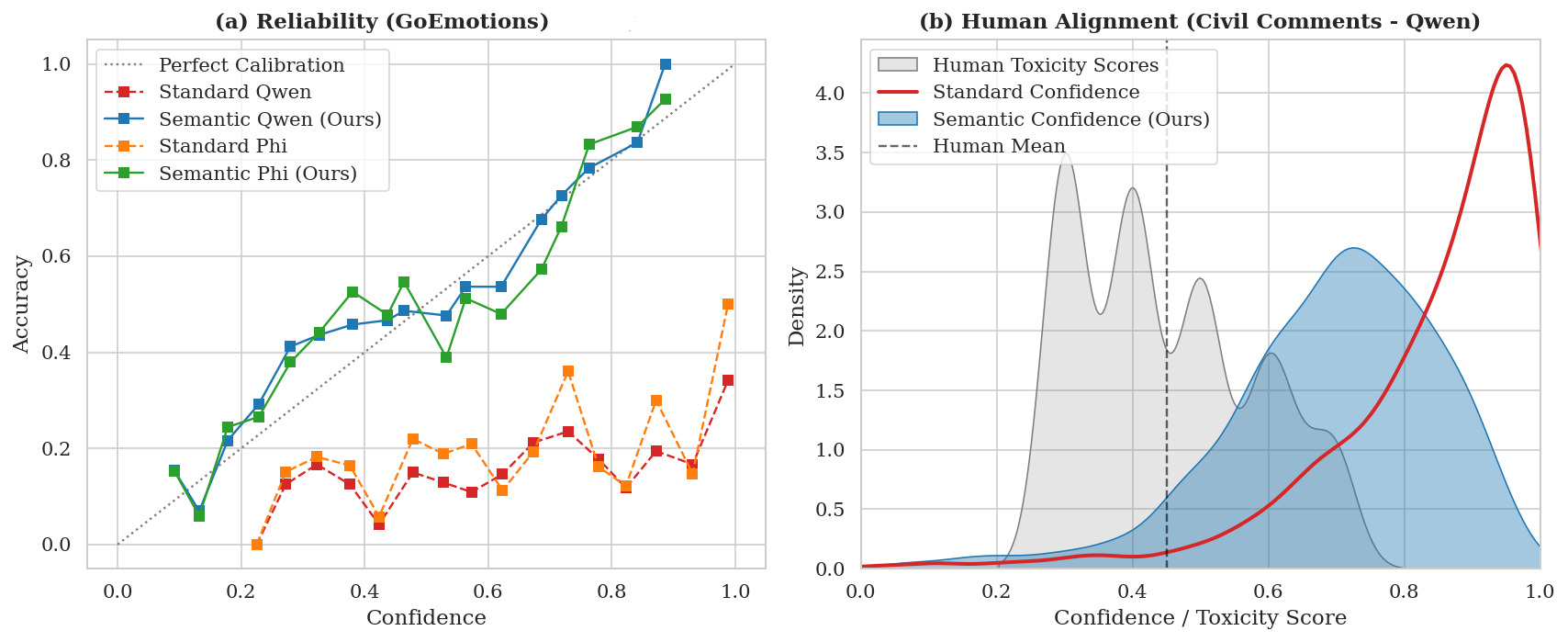} 
    \caption{(a) Reliability: Standard decoding exhibits systematic overconfidence (below diagonal), while Semantic Softmax tracks the ideal calibration curve. (b) Confidence: On ambiguous inputs, standard models force extreme probabilities ($>0.9$), whereas Semantic Softmax aligns with human consensus, effectively capturing aleatoric uncertainty.}
    \label{fig:reliability_dist}
\end{figure*}

\section{Results and Analysis}

We evaluate the efficacy of Semantic Softmax in mitigating Renormalization Bias across our benchmarks. Our evaluation focuses on model calibration, discriminative power, and alignment with human uncertainty.

\subsection{Quantitative Performance and Calibration}

Table \ref{tab:main_results} summarizes the performance metrics across architectures and datasets. The primary finding is a systemic and substantial reduction in ECE. On GoEmotions, Semantic Softmax achieves an ECE of 0.071 for Qwen and 0.065 for Phi, representing up to an 8.5-fold improvement over standard constrained baselines. This trend extends to CivilComments, where Semantic Softmax consistently mitigates the systematic overconfidence of standard decoding, reducing ECE by over 75\% on average. Notably, this improvement in probabilistic reliability does not come at the cost of classification quality; we observe a consistent uplift in discriminative signal, with AUROC and Macro-F1 increasing across all tested configurations. 

This suggests that aggregating the probability mass of semantic neighbors not only improves calibration but also enhances discriminative accuracy by recovering previously unobserved probability mass.

\begin{comment}
\begin{table}[ht]
\centering
\small
\begin{tabular}{llcccc}
\toprule
\textbf{Dataset} & \textbf{Method} & \textbf{ECE $\downarrow$} & \textbf{Brier $\downarrow$} & \textbf{AUROC $\uparrow$} & \textbf{F1 $\uparrow$} \\
\midrule
GoEmotions & Standard & 0.574 & 0.842 & 0.712 & 0.162 \\
& \textbf{Semantic} & \textbf{0.071} & \textbf{0.610} & \textbf{0.725} & \textbf{0.171} \\
\midrule
Civil Ambiguous & Standard & 0.482 & 0.571 & 0.784 & -- \\
& \textbf{Semantic} & \textbf{0.112} & \textbf{0.523} & \textbf{0.791} & -- \\
\bottomrule
\end{tabular}
\caption{Comparison of Standard vs. Semantic Softmax. Results indicate superior probability alignment while maintaining or improving discriminative power (AUROC).}
\label{tab:main_results}
\end{table}
\end{comment}

\subsection{Reliability and Confidence Distribution}

Figure \ref{fig:reliability_dist} illustrates the calibration impact. Panel (a) presents a reliability diagram, where the diagonal identity line ($y=x$) represents perfect calibration—a state where predicted confidence exactly matches empirical accuracy. Deviations falling below this line indicate overconfidence, where the model's probability estimates exceed its actual correctness.

As observed, the standard constrained baseline exhibits a distinct curve significantly below the diagonal, indicative of the systematic overconfidence driven by Renormalization Bias. In contrast, Semantic Softmax minimizes ECE, yielding a reliability curve that closely tracks the ideal diagonal.

This calibration is mirrored in the confidence distribution shift shown in Panel (b). Within the ambiguous Civil Comments subset (human toxicity 0.3--0.7), standard decoding converges to extreme probabilities ($P > 0.90$), effectively erasing annotator disagreement. Semantic Softmax, by aggregating local semantic mass, produces calibrated confidence scores that align with the human mean, accurately quantifying the aleatoric uncertainty inherent in the input.

\section{Conclusion}

In this work, we identified Renormalization Bias as a primary source of systematic overconfidence and poor calibration in zero-shot classification. We demonstrated that restricting decoding to a sparse label set inadvertently discards the probability mass of semantic synonyms, artificially inflating the confidence of the remaining targets.

To address this, we introduced Semantic Softmax, an inference-time intervention that aggregates probability mass from the model's semantic neighborhood. Our experiments confirm that this approach drastically reduces ECE on the GoEmotions and Civil Comments benchmark without compromising discriminative performance. By effectively recovering the lost signal from the full vocabulary, Semantic Softmax allows LLMs to accurately quantify aleatoric uncertainty, offering a robust and calibrated alternative for classification tasks.

\section*{Limitations}

A primary limitation of our current study is its focus on the LLM as a Classifier paradigm. Our method is specifically designed for tasks where models are prompted to select a label from a predefined set of candidates during constrained decoding.

Our empirical validation focuses on efficient, small architectures. Due to computational resource constraints, we restricted our analysis to this regime and have not yet established the scaling laws of Semantic Softmax on large-scale foundation models. 

Semantic Softmax introduces a non-negligible computational cost at inference time. Retrieving the semantic neighborhood requires a dynamic top-k search and similarity calculation over the vocabulary space. While negligible for single-token classification, this overhead prevents the technique from being easily applied to long-form decoding tasks where latency is cumulative.

Furthermore, the method's efficacy is intrinsically bound to the quality of the underlying embedding space. Our approach assumes that semantically related tokens reside in close geometric proximity. In models with highly anisotropic embeddings or poor semantic clustering, aggregating neighbors could theoretically introduce noise rather than signal. Finally, our evaluation is currently limited to English-language benchmarks . We leave the exploration of Semantic Softmax in multilingual settings where synonyms may span across languages and token to future work.

\section{Ethics Statement} 
While Semantic Softmax successfully mitigates Renormalization Bias and improves confidence calibration in sensitive domains like toxicity and emotion classification, it does not eliminate the societal biases embedded within the base models' embedding spaces. We emphasize that accurately quantified uncertainty signals linguistic ambiguity, not objective fairness. Therefore, we strongly advise practitioners deploying this methodology in public moderation or behavioral analysis systems to use high-uncertainty outputs as a trigger for human-in-the-loop review, ensuring responsible handling of linguistic nuances rather than relying solely on automated automated judgments.

\bibliography{custom}

\appendix
\section{Appendix}
\label{sec:appendix}

\subsection{Qualitative Analysis of Renormalization Bias}
To illustrate the impact of Renormalization Bias, we provide specific examples from the Civil Comments dataset in Table~\ref{tab:qualitative_results}. These samples demonstrate how standard constrained decoding tends toward extreme overconfidence, while Semantic Softmax recovers the Silent Vote to align more closely with human uncertainty.

\begin{table*}[ht]
\centering
\small
\begin{tabularx}{\textwidth}{X c c c}
\toprule
\textbf{Text Sample} & \textbf{Human Truth} & \textbf{Std Score} & \textbf{Sem Score} \\
\midrule
Let thouse who curse our flag repent or be cursed. & 0.478 & 0.766 & 0.599 \\
Investing in infrastructure is one thing... The best thing that could happen is to stop the idiocy of the emx boondoggles. & 0.600 & 0.906 & 0.661 \\
I can't be expected to rebut *every* dopey comment made here and at the National Post, CFL. & 0.400 & 0.664 & 0.452 \\
They need to euthanize this animal before he kills again. & 0.700 & 0.910 & 0.762 \\
OK Then if it's so common... Put up or shut up. & 0.700 & 0.957 & 0.822 \\
Looks like the Judge might have gotten some "side benefits"... throw them in jail. & 0.500 & 0.652 & 0.576 \\
Good defense strategy: "We wuz too dumb to do anyting bad." & 0.600 & 0.906 & 0.697 \\
He's a Hispanic causing a panic. & 0.600 & 0.789 & 0.727 \\
You're using someone else's name, you're a troll. & 0.667 & 0.938 & 0.879 \\
Perhaps not everyone is as grotesquely course as you! & 0.500 & 0.547 & 0.590 \\
So, this would be the alternative to the nuclear deal dressed up as a pigs ear? Oops, sorry, silk purse... & 0.600 & 0.816 & 0.723 \\
Let the NFL die. It's what it deserves. & 0.600 & 0.910 & 0.797 \\
Good response, JT; I enjoy your even-handed participation... And thank you for your military service. & 0.167 & 0.295 & 0.200 \\
Fox News is the meaning of nitwit. & 0.700 & 0.926 & 0.746 \\
CNN haha can dish but can't take. Losers. Go away. & 0.522 & 0.910 & 0.636 \\
Gosh Justin, this sure makes you appear corrupt. Take a picture of that. & 0.500 & 0.797 & 0.603 \\
Oh, a cross country tour - sounds ever so nice. What a pompous twit he is! & 0.613 & 0.926 & 0.718 \\
I would just mute this troll. & 0.700 & 0.941 & 0.716 \\
What do you do with an elephant with three balls? & 0.338 & 0.053 & 0.207 \\
Another frustrated juvenile response, naked. & 0.400 & 0.621 & 0.439 \\
\bottomrule
\end{tabularx}
\caption{Qualitative comparison of toxicity predictions. Semantic Softmax (Sem Score) mitigates extreme probabilities by aggregating local semantic mass, tracking the Human Truth (toxicity mean) more accurately than standard constrained decoding (Std Score).}
\label{tab:qualitative_results}
\end{table*}

\subsection{Hyperparameter Sensitivity}
The Semantic Softmax framework relies on two primary hyperparameters: the number of top vocabulary tokens $K$ and the noise filter threshold $\tau$. We found that performance is robust to $K$ values between 100 and 1000, provided the semantic kernel is properly thresholded. Specifically, as shown in Table~\ref{tab:qwen_sensitivity}, we identified $\tau = 0.80$ as the optimal noise filter threshold for minimizing ECE.

\begin{table*}[ht]
\centering
\small
\setlength{\tabcolsep}{3pt} % Reduces space between columns to prevent overlap
\begin{tabularx}{\textwidth}{@{} l *{6}{>{\centering\arraybackslash}X} @{}}
\toprule
\multicolumn{7}{c}{\textbf{Table A: Expected Calibration Error (ECE)} $\downarrow$} \\
\midrule
$K \downarrow / \tau \rightarrow$ & \textbf{0.70} & \textbf{0.75} & \textbf{0.80} & \textbf{0.85} & \textbf{0.90} & \textbf{0.95} \\
\midrule
\textbf{50} & 0.1532 & 0.1497 & \textbf{0.1145} & 0.1840 & 0.1839 & 0.1840 \\
\textbf{100} & 0.1523 & 0.1514 & \textbf{0.1145} & 0.1923 & 0.1923 & 0.1923 \\
\textbf{200} & 0.1540 & 0.1492 & 0.1168 & 0.1987 & 0.1986 & 0.1986 \\
\textbf{300} & 0.1538 & 0.1491 & 0.1144 & 0.1987 & 0.1989 & 0.1987 \\
\textbf{400} & 0.1529 & 0.1511 & 0.1154 & 0.1986 & 0.1986 & 0.1987 \\
\textbf{500} & 0.1528 & 0.1489 & 0.1142 & 0.1988 & 0.1986 & 0.1987 \\
\textbf{600} & 0.1527 & 0.1498 & 0.1130 & 0.1986 & 0.1985 & 0.1986 \\
\textbf{700} & 0.1538 & 0.1487 & 0.1141 & 0.1986 & 0.1986 & 0.1987 \\
\textbf{800} & 0.1527 & 0.1498 & 0.1151 & 0.1995 & 0.1994 & 0.1995 \\
\textbf{900} & 0.1517 & 0.1508 & 0.1151 & 0.1993 & 0.1992 & 0.1994 \\
\textbf{1000} & 0.1527 & 0.1499 & 0.1162 & 0.1993 & 0.1992 & 0.1993 \\
\midrule
\multicolumn{7}{c}{\textbf{Table B: Macro-F1 Score} $\uparrow$} \\
\midrule
$K \downarrow / \tau \rightarrow$ & \textbf{0.70} & \textbf{0.75} & \textbf{0.80} & \textbf{0.85} & \textbf{0.90} & \textbf{0.95} \\
\midrule
\textbf{50} & 0.4092 & 0.4101 & 0.4325 & \textbf{0.4359} & 0.4359 & 0.4359 \\
\textbf{100} & 0.4083 & 0.4102 & \textbf{0.4308} & 0.4112 & 0.4112 & 0.4112 \\
\textbf{200} & 0.4083 & 0.4093 & \textbf{0.4308} & 0.3958 & 0.3958 & 0.3958 \\
\textbf{300} & 0.4083 & 0.4093 & \textbf{0.4308} & 0.3909 & 0.3909 & 0.3909 \\
\textbf{400} & 0.4083 & 0.4093 & \textbf{0.4308} & 0.3899 & 0.3899 & 0.3899 \\
\textbf{500} & 0.4083 & 0.4093 & \textbf{0.4308} & 0.3899 & 0.3899 & 0.3899 \\
\textbf{600} & 0.4083 & 0.4093 & \textbf{0.4308} & 0.3899 & 0.3899 & 0.3899 \\
\textbf{700} & 0.4083 & 0.4093 & \textbf{0.4308} & 0.3899 & 0.3899 & 0.3899 \\
\textbf{800} & 0.4083 & 0.4093 & \textbf{0.4308} & 0.3889 & 0.3889 & 0.3889 \\
\textbf{900} & 0.4083 & 0.4093 & \textbf{0.4308} & 0.3889 & 0.3889 & 0.3889 \\
\textbf{1000} & 0.4074 & 0.4093 & 0.4317 & 0.3889 & 0.3889 & 0.3889 \\
\bottomrule
\end{tabularx}
\caption{Hyperparameter sensitivity analysis. We find that $\tau = 0.80$ consistently minimizes ECE while maintaining high Macro-F1 across a wide range of $K$, effectively mitigating the \textit{Renormalization Bias}.}
\label{tab:qwen_sensitivity}
\end{table*}

\end{document}